\documentclass[10pt, conference]{IEEEtran}
\IEEEoverridecommandlockouts
\usepackage{cite}
\usepackage{amsmath,amssymb,amsfonts}
\usepackage{algorithmic}
\usepackage{graphicx}
\usepackage{subfig}
\usepackage{textcomp}
\usepackage{multirow}
\usepackage{xcolor}
\def\BibTeX{{\rm B\kern-.05em{\sc i\kern-.025em b}\kern-.08em
    T\kern-.1667em\lower.7ex\hbox{E}\kern-.125emX}}

\usepackage[english]{babel}

\usepackage{amsmath}
\usepackage{graphicx}
\usepackage[colorlinks=true, allcolors=blue]{hyperref}
\usepackage{amsfonts}

\usepackage[colorinlistoftodos]{todonotes}

\renewcommand*\descriptionlabel[1]{\hspace\leftmargin$#1$}

\title{\textit{Storehouse}: a Reinforcement Learning Environment for Optimizing Warehouse Management}

\author{\IEEEauthorblockN{Julen Cestero}
\IEEEauthorblockA{\textit{Data Intelligence for Energy and Industrial Processes} \\
\textit{Vicomtech}\\
San Sebastian, Spain \\
0000-0002-6670-6255}
\and
\IEEEauthorblockN{Marco Quartulli}
\IEEEauthorblockA{\textit{Data Intelligence for Energy and Industrial Processes} \\
\textit{Vicomtech}\\
San Sebastian, Spain \\
0000-0001-5735-2072}
\and
\IEEEauthorblockN{Alberto Maria Metelli}
\IEEEauthorblockA{\textit{Dipartimento di Elettronica, Informazione e Bioingegneria} \\
\textit{Politecnico di Milano}\\
Milano, Italy \\
0000-0002-3424-5212}
\and
\IEEEauthorblockN{Marcello Restelli}
\IEEEauthorblockA{\textit{Dipartimento di Elettronica, Informazione e Bioingegneria} \\
\textit{Politecnico di Milano}\\
Milano, Italy \\
0000-0002-6322-1076}
}

\begin{document}
\maketitle

\begin{abstract}
    Warehouse Management Systems have been evolving and improving thanks to new Data Intelligence techniques. However, many current optimizations have been applied to specific cases or are in great need of manual interaction. Here is where Reinforcement Learning techniques come into play, providing automatization and adaptability to current optimization policies. In this paper, we present \textit{Storehouse}, a customizable environment that generalizes the definition of warehouse simulations for Reinforcement Learning. We also validate this environment against state-of-the-art reinforcement learning algorithms and compare these results to human and random policies.
\end{abstract}

\section{Introduction}




    Advancements in Warehouse Management Systems can be classified into different trends. On the one hand, some advancements aim to enhance specific processes of the warehouses, improving the effectiveness and efficiency of the whole activity. On the other hand, other advances focus on  improving more general processes, using  \emph{Machine Learning} methods such as \emph{Reinforcement Learning}.

    {Reinforcement Learning} (RL)~\cite{sutton2018reinforcement} is a machine learning approach for training agents to solve sequential decision-making problems by interacting in a trial-and-error fashion \cite{OpenAI2018}. RL is an iterative approach in which an agent receives rewards based on the actions taken, with the objective of maximizing the cumulative reward during its experience in the environment. One of the advantages of RL is that, given a well-defined environment and reward, it can progressively adapt to new arrangements within the envlironment to continue maximizing the cumulative reward. Furthermore, RL is not constrained to a specific topology or dynamics of the environment, being able to generalize across different configurations. Therefore, RL turns out to be a potential solution to the problem of managing warehouse storage systems.

    In this paper, we investigate RL approaches to tackle warehouse storage management. The main contributions of the paper can be summarized as follows:
    \begin{itemize}
    \item we formulate warehouse storage management as a \emph{Markov Decision Process} (MDP)~\cite{puterman1994markov}. The modeling is general and can be customized with different warehouse topologies and generation processes (Section~\ref{sec:problem});
    \item we introduce \emph{Storehouse}, a customizable environment designed to train RL agents to manage the storage of a warehouse using a FIFO methodology. The environment is implemented using the \emph{gym} interface \cite{OpenAIgym} to ease connection to popular RL libraries such as stable-baselines3 \cite{stable-baselines3} or RLlib \cite{Rllib}. {Storehouse} also includes a customization method that uses a configuration file, where developers can try out different versions of the environment, including different grid sizes, material types, and timers (Section~\ref{sec:solution});
    \item we provide an experimental evaluation of several RL algorithms, including DQN~\cite{Mnih2016}, PPO~\cite{schulman2017proximal}, A2C~\cite{Mnih2016}, against handcrafted human policies, showing competitive performance (Section~\ref{sec:results}).\footnote{The code of the environment is available in Github (\url{https://github.com/JulenCestero/storehouse}).}
    \end{itemize}
    
\section{Current Methodologies}
    Warehouse management systems follow different methodologies to optimize and balance their workload. Apart from a few very recent facilities, such as Amazon's \textit{Kiva} system \cite{Yang2020}, most modern warehouses follow classic approaches. However, with the rise of Industry 4.0, many warehouses have started looking for more data-driven solutions to optimize their long-established processes, which usually depend on the experience of senior managers and \textit{ad hoc} methods.

    These methods are largely explored in the literature and can be grouped into two trends: a \emph{classic} trend that aims to enhance current systems via optimization (Section~\ref{sec:classicTrends}), and a more \emph{modern} trend, which aims to propose methods, breaking with existing ones (Section~\ref{sec:modernTrends}). 
    
\subsection{Classic Trends}\label{sec:classicTrends}
    The classic trend leverages advancements in the area of sensors and automation, to enhance existing warehouse \textit{ad hoc} methods using real data instead of human experience \cite{Zunic2018}. These improvements rely on location sensors, product identification tags, motion controllers. Using these methodologies, several authors propose different optimization methods. Liang et al. \cite{Liang2020} propose a wave-picking warehouse management that shows a 2\% improvement compared with a baseline item-batching method. Mao et al. \cite{Mao2018} provide a general study of the IT systems in a warehouse and propose a scheduling function aimed at the interconnection of several agents in an industrial management process, which involves several warehouses, a fleet of vehicles, and some manufacturers and consumers. Zunic et al. \cite{Zunic2018,Zunic20182} propose two upgrades to the logistics of a particular warehouse. The first one consists in managing the distribution pipeline of supplies and orders of the warehouse, while the second one uses an inner optimization of the ordering and distribution of the material on the warehouse shelves. Both have some promising results that improve the previous implementation close to 12\%.
    Finally, Zunic et al. \cite{Zunic2017} analyze a material reallocation of the warehouse using an algorithm that improves previous efficiency by 25 times. 
    
    All these methodologies share the same limitations: they are aimed at solving specific problems of certain warehouses, they do not provide a general methodology that can be applied to any kind of management. These improvements can be useful for micro management optimizations, but we aim to provide more general management methodologies that can be adaptable to many kinds of warehouses.

\subsection{Modern Trends}\label{sec:modernTrends}
   More recently, approaches are based on \emph{machine learning} and  complex \emph{data analysis} have been proposed to design alternatives to classic techniques, taking advantage of the aforementioned new developments on warehouse digitalization.
   Among all of them, a relevant role is played by RL.
   These methods can be classified according to which portion of the process they optimize: the \emph{macro} actions, planning of higher order actions, like, for example, the ordering of new upcoming materials to the warehouse; and the \emph{micro} actions, which are based on the specific movements of the agents inside the warehouse. This latter kind of optimization focuses on the path-finding process needed to find the optimal path to fulfill the macro decisions. Bottani et al. \cite{Bottani2015} carried out a very detailed review analyzing the most time-consuming activity in a warehouse and providing an exhaustive research encompassing several state-of-the-art types of algorithms, such as \emph{genetic algorithms}, ant colony optimization, particle swarm optimization, simulated annealing, and methods that use \emph{artificial neural networks}. Estanjini et al. \cite{Estanjini2012} propose an RL-based method using A2C \cite{Mnih2016} to provide high-level actions in the decision-making process, such as moving items to/from reserve/pick-up/deport locations.
   They divide their action space into these macro-activities, address the curse of dimensionality \cite{Bellman1961} by decreasing the number of action-state combinations, and apply the algorithm managing to improve performance by 20\% compared to a similar heuristic previously in use. However, this approach is limited to the specific layout of the studied warehouse. Dou et al. \cite{Dou2015} proposed a mixed method involving {Genetic Algorithms} (GAs) with RL. GAs are used for the planning part of the process, involving the macro-action management, while the RL algorithms are used for the micro-actions of the agents that involve instantaneous movement and path-finding throughout the grid. Thus, the GA defines the goal of the agent, and the RL algorithm manages the movements to achieve that goal. The results of this article provide insight that has been verified by other authors \cite{Li2019, Yang2020, Lee2021}. These works compare RL algorithms such as Q-Learning \cite{Watkins1992} or DQN \cite{Mnih2016} with classic path-finding algorithms, such as A* \cite{Hart1968}, and conclude that RL algorithms achieve at least the same efficiency as A*, and enhance it in some cases, but all these works are limited to address the path-finding part of the warehouse management problem, without addressing the macro-planning problem using complex data analysis.

  
  The main limitation of these approaches is that none of them proposed a generalized environment. A generalized environment can be used to train RL agents and, therefore, propose a global method that sums up both the micro and the macro management of a real warehouse.


\section{The Storehouse Problem}\label{sec:problem}
Although advances in Industry 4.0 have provided new tools for warehouses optimization, there is still a strong need to automate these improvements and making them adaptable to new and unexpected scenarios. Nevertheless, much of the management is highly dependent on human interaction and, in many cases, leading to \textit{ad hoc} methods originating from cumulative human experience. In this sense, artificial intelligence approaches, and in particular RL, can greatly improve warehouse management by devising approaches that are responsive and adaptable to changes inside the warehouse. In this section, we first provide an informal description of the \emph{Storehouse} problem by discussing the \emph{metrics} employed to evaluate the performance of the warehouse management (Section~\ref{sec:metrics}) and the general warehouse features (Section~\ref{sec:envDescr}). Then, according to the previous description, we provide the environment specification as MDP (Section~\ref{sec:mdp}). 

The goal of the \emph{Storehouse} problem is to micro-manage a number of autonomous agents (e.g., fork lifts), providing them with actions or movements to accomplish. Therefore, we seek to design a centralized control unit to manage the logistics of the warehouse, optimizing the metrics discussed below.
    
\subsection{Metrics}\label{sec:metrics}
In the literature, several metrics are available to evaluate the performance of a warehouse management system. However, in this paper, we mainly focus on three of them: 
\begin{enumerate}
	\item[(i)] the \emph{number} of material or orders delivered in a defined time period;
	\item[(ii)] the \emph{average age} of the material in the warehouse over a time period;
	\item[(iii)] the number of times the FIFO criterion has being \emph{violated} (the FIFO criterion is widely used in current warehouse management systems).
\end{enumerate}

The choice of these metrics is justified by the use case of interest. In particular, we focus on the number of materials or orders delivered due to the nature of the real warehouse, which provides industrially manufactured products. If the products stored in the real warehouse were perishable goods, for instance, other metrics would certainly be more important, as the age of the material. 
    

\subsection{Warehouse Description}\label{sec:envDescr}
In this section, we provide a description of the abstraction of the warehouse functioning that we consider in this paper.

\subsubsection{Warehouse Layout}
\begin{figure}[t]
        \centering
        \includegraphics[width=.28\textwidth]{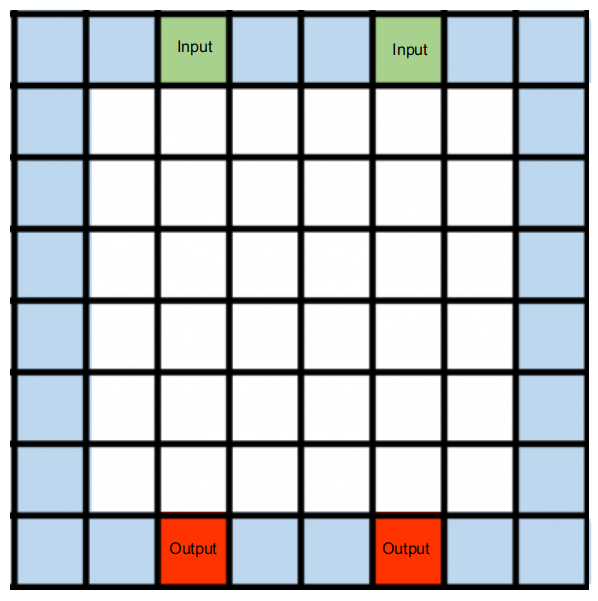}
        \caption{\label{fig:grid}Layout of the warehouse. The blue tiles correspond to the area that is always free of material. The entry points are green, while the delivery points are red.}
    \end{figure}

The warehouse layout is a rectangular grid with a hollow crown that surrounds the storage room to ease handling of forklifts. An example of a squared layout is shown in Figure~\ref{fig:grid}. The warehouse contains two \emph{entry} points (green) and two \emph{delivery} points (red), where items are introduced and delivered to clients. These points are located in an outer crown where no items can be placed with the exception of the delivery points, which immediately consume the items placed therein. Therefore, the items can be placed \emph{anywhere} within the internal storage grid. We assume the size of the material is equal to one grid tile. Items placed in the storage can be picked if the agent is able to get to that cell from adjacent horizontal or vertical cells. If a cell is unreachable, this cell is considered a \emph{restricted cell}, where no items can be placed or picked. 

\subsubsection{Item Generation and Consumption}
The item generation process starts at the delivery points. By emulating \textit{on-demand} material generation systems, they generate the orders, using random timers, which the system has to fulfill. The entry points start generating the required items, placing them into waiting queues with timers that simulate when the different materials of the order are ready to be either introduced into the grid or delivered directly to the delivery points.
In our setting, timers are Poisson processes with a configurable parameter $\lambda$~\cite{wolff1982poisson}. In summary, the delivery points generate a new order with a random distribution, and a random timer with the waiting time until ready to collect the order; this order is passed to the entry points and they start creating the material in a queue, with random timers simulating it. When the timers of the items of the input queues finish, the items are ready to be picked up and moved to the storage or delivered directly to the delivery points, and they are collected from the input points in the order they were generated.

\subsubsection{Material Types}
Different types of material are considered to mimic actual production and delivery to clients, named in alphabetical order (e.g., the first material type is `A', the second `B', and so on). The generation is governed by different stochastic processes (e.g., Poisson processes with different $\lambda$s), and the order type is also randomly defined by the delivery points when a new order is created.

    \subsection{MDP Specification} \label{sec:mdp}
Based on the informal description presented above, in this section, we formally define all the elements needed for modeling the problem as an MDP~\cite{puterman1994markov}.
        
        \subsubsection{States}

        The state of the environment is composed of several grids, each storing a feature of all the cells in the warehouse, easing the training by making it compatible with image processing methods. 
        Formally, the state is defined as a 3-dimensional tensor $s \in \mathbb{R}^{r\times c\times d}$, where $r,c \in \mathbb{N}$ are the number of rows and columns of the warehouse grid respectively and $d \in \mathbb{N}$ is the number of feature matrices of the state. Thus, $s$ can be regarded as a list of $r \times c$ feature matrices $D_i$. We consider $d=6+m$ feature matrices, where $m$ is the number of material types considered:\footnote{For compliance with convolutional neural networks, all matrices are treated as images and their values are normalized between $0$ and $255$.}


        
        \begin{itemize}
            \item \emph{Box grid} $D_1$: represents the location of the different boxes within the storage. Each box is represented by an integer corresponding to the type of material it is, mapping each type (`A', `B', ...) to integers (1, 2, ...). If there is no box in that grid position, the value is 0.
            \item \emph{Age grid} $D_2$: represents the age of the box, which is bounded between 0 and 1000 (steps).
            \item \emph{Restricted grid} $D_3$: collects the different restricted cells in the environment. 
            \item \emph{Agent grid} $D_4$: 
            used to locate the agent within the grid and to know if the agent has picked an object or not. The cell in the matrix in which the agent is placed will have a value of 255 if the agent has not picked an object, and 128 otherwise. The other cells will have a value of 0.
            \item \emph{Agent material grid} $D_5$: 
            the cell where the agent is located is marked with the type of the material that it is carrying is represented. If the agent has no items, the value of all cells is 0.
            \item \emph{Entry point grid} $D_6$:
             indicates the status of the entry points. When an item is ready to be picked up, the position of the matrix that corresponds to the position of the corresponding entry point changes its value to 255.
            \item \emph{Out point grids}: $D_7,\dots,D_{6+m}$: the number of matrices varies with the number of types that the environment can generate, each matrix corresponding to a different type. Depending on the type of material that the delivery points are expecting, the value of the cell at the position corresponding to the delivery points changes to 255 in the corresponding matrix.
        \end{itemize}



\subsubsection{Actions}
       The available actions take the form of coordinates of the grid $a = (i,j) \in \{1,\dots,r\} \times \{1,\dots,c\} $. 
        Any action corresponds to the \emph{desired location} of the agent.
        The effect of the action depends on the status of the target cell:
        \begin{itemize}
        	\item if the agent moves to the same location where a box is located, it will automatically pick it up;
        	\item if the agent has an object and it moves to an empty valid cell, it will automatically drop the item there.
        \end{itemize}
%
         However, some actions are considered invalid. For instance, the agent cannot stack items or deposit an item in the outer ring of the environment, unless it is depositing it at an open delivery point. If an invalid action is performed, the agent will receive a negative reward and it will stay in the same cell as before, even though the environment will advance one step further without registering any movement from the agent. Given a state $s$, these actions belong to the set $\mathcal{I}_{\text{nv}}(s)$.

\subsubsection{Reward}
We assume that the cost of a movement is constant (one step), regardless of the distance involved in the movement. 
These actions result in a reward that values the quality of the outcome of the performed action:
        
        \begin{itemize}
            \item if the agent performs an invalid action, the reward will always be $-1$, with the objective of preventing the agent from performing an invalid action;
            \item when the agent has nothing useful to do, e.g., when the grid is empty and there are no incoming orders, it will receive a reward equal to $0$, preventing actions from having a significant effect in the training;
            \item if the agent has other significant actions to take (e.g., delivering ready items) and does not perform them, it will receive a reward of $-0.9$ to discourage idle actions;
            \item if the agent manages to deliver an item to the delivery points, it will receive a less negative reward, to encourage it to this behavior. This reward is bounded between $[-0.5, 0]$ and it will be closer to $0$ the older the delivered box is, with the maximum reward being $0$ if the box is the oldest available and the agent is therefore behaving according to a FIFO criterion. The reward will vary linearly between the upper and lower bounds, depending on the age difference between the delivered box and the youngest box in the environment and, additionally, this age difference is also bounded between $0$ and $100$ steps, returning the highest and the lowest rewards respectively.
        \end{itemize}



        We can formalize the definition of the rewards for every state-action pair $(s,a)$:
        \begin{equation*}
            R(s,a) = \begin{cases}
                -1      & a \in \mathcal{I}_{\text{nv}}(s)\\
                -0.9    & a \in \mathcal{N}(s) \\
                \frac{\min \{\max \{\text{age}, 0\}, 1000\}}{1000} \cdot 255 & a \in \mathcal{D}(s) \\
                0       & \text{otherwise}
        \end{cases},
        \end{equation*}
        where:
        \begin{itemize}
            \item $ \mathcal{I}_{\text{nv}}(s)$ is the set of invalid actions in state $s$;
            \item $\mathcal{D}(s)$ is the set of valid actions that deliver items to the delivery points in state $s$;
            \item $\mathcal{N}(s)$, is the set of idle actions in state $s$ (i.e., the agent moving from an empty cell to another with no items, with waiting items at the entry points).
        \end{itemize}

        There is no special reward for items collected from entry points. If there are items to be delivered, the agent receives $-0.9$ reward for avoiding idleness. However, if there are no more effective actions than getting new items, the reward is $0$.

        In this setting, the goal of an RL agent is to find the policy $\pi$, i.e., a mapping from states to actions, that maximizes the expected discounted sum of the rewards~\cite{sutton2018reinforcement}:
        \begin{equation*}
            J(\pi) = \mathbb{E}_{\pi} \left[\sum_{t=0}^{T-1} \gamma^t R(s_t, a_t) \right],
        \end{equation*}
where $\gamma \in (0, 1)$ is the discount factor.

\section{Algorithmic Solution}\label{sec:solution}

    To solve the \textit{Storehouse} optimization problem, we use RL-based strategies and algorithms. Our goal is to show that RL solutions are competitive with traditional human-knowledge-based strategies.
%



\subsection{Reinforcement Learning Policies}
    We tested three state-of-the-art RL algorithms in the \emph{Storehouse} environment: \emph{Deep Q-Network} (DQN)~\cite{Mnih2016}, \emph{Asynchronous Actor Critic} (A2C)~\cite{Mnih2016}, and \emph{Proximal Policy Optimization} (PPO)~\cite{schulman2017proximal}. In addition, we developed a customized version of DQN, in which we filter out the invalid actions thus speeding up the training process due to the decreased number of actions available in each step. We called this variant \textit{Valid-Action Mask} (VAM). 

    \subsection{Hyperparameter Definition}\label{sec:hyperp}
    We also studied the effect of hyperparameter optimization, compared to using the default values defined in stable-baselines3, the library we used for the RL training. 
    In this study, we sought strategies based on Design of Experiments (DoE) \cite{DoE} to optimize the training of the RL policies. 
    For this purpose, we quantized the continuous hyperparameters and applied a pairwise combination, used in all-pairs testing \cite{czerwonka2006pairwise}, to reduce the space of possible combinations. 
     This optimization allows increasing the speed of convergence to the optimal combination of hyperparameters.


    We considered four main hyperparameters, from the ones that stable-baselines3 allows us to tune:
    \begin{itemize}
        \item \textit{learning rate} ($\alpha$): 
        choosing the learning rate is challenging as too small a value may result in a long training process that could get stuck, whereas too large a value may result in learning a sub-optimal set of weights too fast or an unstable training process;
        \item \textit{discount factor} ($\gamma \in (0,1)$): it quantifies the importance we give to future rewards. If $\gamma$ is closer to $0$, the agent will tend to consider only immediate rewards. If $\gamma$ is closer to $1$, the agent will give similar weights to immediate and future rewards, willing to delay the reward;
        \item \textit{Polyak coefficient} ($\tau \in [0,1]$): it is the soft update coefficient, being $1$ a hard update. It averages previous values of the neural network with the most recent values;
        \item \textit{exploration fraction}: it is the fractional value for the exploration vs. exploitation rate, in this case, an $\varepsilon$-greedy strategy. It represents the percentage of steps of the defined maximum number of training steps that the algorithm needs to reach the minimum value of $\epsilon$. For instance, a value of $0.1$ means that the value of $\epsilon$ will vary linearly from its maximum value to its minimum in the initial $10\%$ of the defined maximum number of steps.
        \item \textit{Value Function coefficient} (\textit{vf\_coef} $\in [0, 1]$): this coefficient controls the effect of the value function loss in the calculation of the training loss;
        \item \textit{clip} ($\backepsilon~\in [0, 1]$): the standard PPO algorithm has a clipped objective function and this hyperparameter is used to define the range in which the probability ratio term is clipped, meaning that the objective function takes the minimum between the original ratio and the clipped ratio, being the clip range $[1 - \backepsilon, 1 + \backepsilon]$.
    \end{itemize}

\section{Results}\label{sec:results}

    In this section, we present the results of the methods proposed in the previous sections in terms of the RL training score, which relates to the discounted cumulative rewards, and of the different metrics for each defined policy. Furthermore, the effect of hyperparameter tuning is studied for all the RL policies, but a deeper analysis is shown for the DQN policy.

\subsection{Baseline Policies}
    We defined two \emph{human} policies, i.e., policies defined programmatically that implement expert strategies applicable in warehouses.
    The first one, named \emph{Initial Human Policy} (IHP), prioritizes retrieving the material from the entry point queues, whereas the second one, named \emph{Enhanced Human Policy} (EHP), prefers to deliver the available material to the delivery points. 
    Both policies work in such a way that, when a new object is generated at the entry points, this is immediately taken when the warehouse is empty.
    IHP will always store these items in the warehouse, but EHP will store the items only if the agent is unable to deliver them directly. When the storage contains items, IHP will act likewise: prioritizing stocking items from the entry points and delivering available items using a FIFO order, when no items are waiting in the entry point queues. On the other hand, EHP does not prioritize the absence of objects at the entry points. If the agent can deliver the items (thus, the delivery points are open for some items) and these items are in the storage, it will prioritize the delivery of items following the FIFO order and, therefore, delivering the oldest available items. This means that the priority is to deliver the items already stored, try to empty the warehouse, and, then, deliver or store the new incoming material from the entry point queues. Besides that, the delivery points do not expect any item, this agent will aim to store the incoming items into the warehouse storage. Furthermore, restricted cells can be created if the incoming flow of new items is greater than the output flow of delivered items. If so, these two policies get the oldest available material in the grid.
     

    We also consider a \emph{random} policy, which chooses uniformly among all actions defined in the action space of the environment at each step. We expect this policy to be the worst, so, we want to use it as a baseline from which the RL policies are expected to improve. The human policies are also considered baselines and we aim for RL policies to outperform the human policies if possible.

\subsection{Experimental Setting}
    All results use the \textit{$6\times 6$} configuration of the environment ($r=c=6$), where entry points in the upper corners and delivery points in the lower corners. We consider two types of material, `A' and `B', with a Poisson generation process with parameters $\lambda = 5$ and $\lambda = 10$ respectively for each item and an order delivery parameter of $\lambda = 30$ and $\lambda = 50$ respectively. Each order has a random number of items between $2$ and $6$, and a new order is randomly generated with $\lambda = 25$.

    The results were obtained by simulating agents with different policies in the \textit{Storehouse} environment, using a maximum number of steps of $1000$ per episode, in $3000$ episodes using a GPU Nvidia Tesla V100 SXM2 32GB, 80 CPUs Intel Xeon Gold 6230 @$2.16$GHz and 755GB of RAM, which results into a duration of nearly $3.5$ hours per trained policy. Each policy is trained 10 times and, in all figures, the results show the average of the different runs with a darker line, and the maximum and minimum over the runs with a shaded area.

    \begin{figure}[t]
        \centering
        \includegraphics[width=0.487\textwidth]{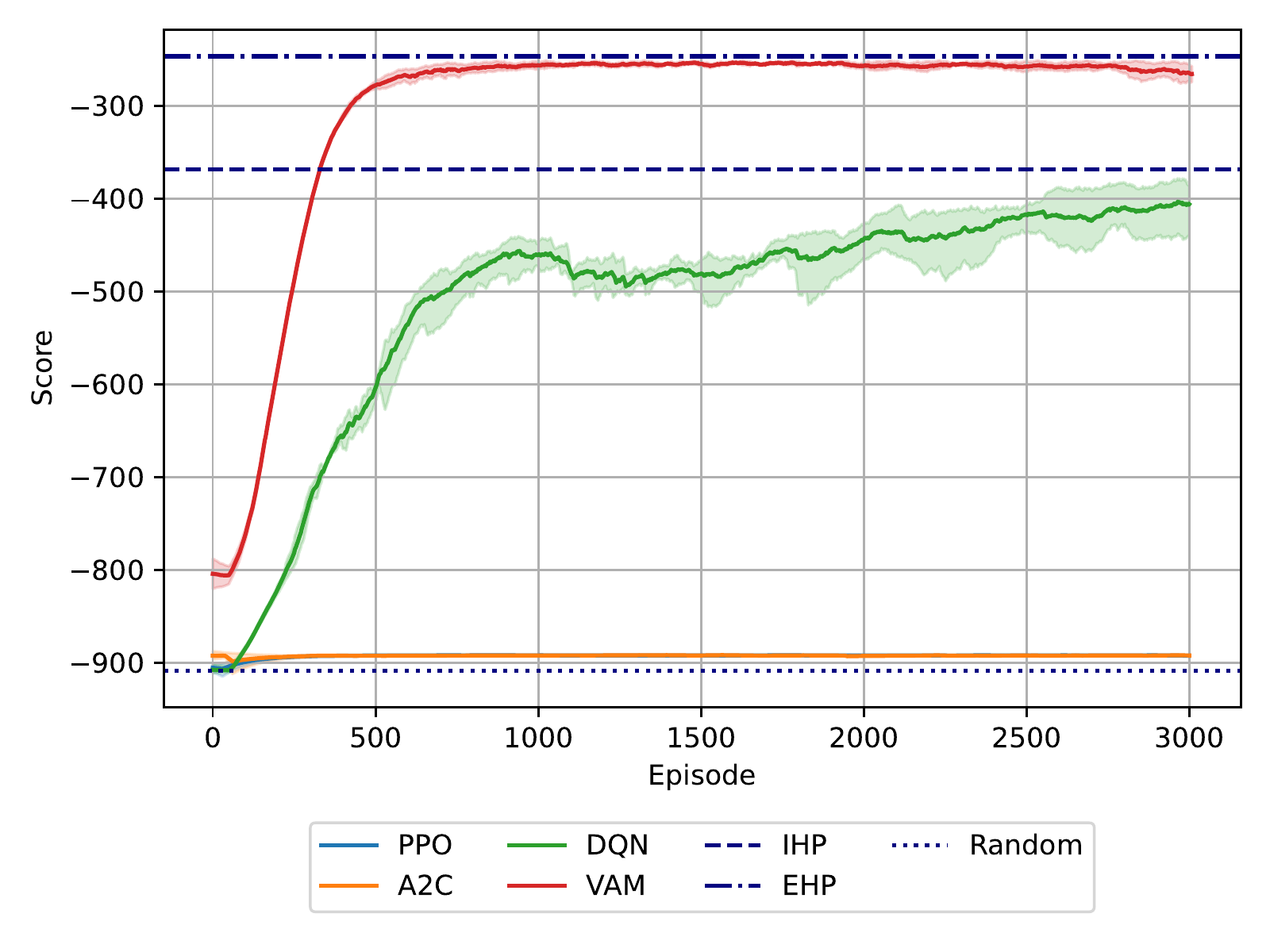}
        \caption{Score as a function of the number of collected episodes for the tested RL algorithms and baseline policies without hyperparameter optimization over 10 runs (PPO line is behind the A2C line). The shaded area corresponds to the area between the maximum and the minimum value between the different runs.}\label{fig:un_algo}
    \end{figure}

\subsection{Results without Hyperparameter Optimization}
    Figure \ref{fig:un_algo} shows the evolution of the average return in the training process per episode for the different defined policies. The RL policies change throughout the episodes, even though some of them stay quite static, but we consider both the human policies and the random policy as constants because they do not vary with the episodes. All the policies in this plot have been trained using the default parameters of their corresponding algorithms in stable-baslines3. It is curious that the policies based on DQN evolve successfully, enhancing their result until achieving near-human performance, but the A2C and PPO policies are not able to learn how to act in the environment. However, it is not surprising that the VAM policy outperforms the DQN policy, even enhancing the initial human policy, since it is an improved version of the default DQN algorithms where the invalid actions are suppressed.

    \begin{figure}[t]
        \centering
        \includegraphics[width=0.487\textwidth]{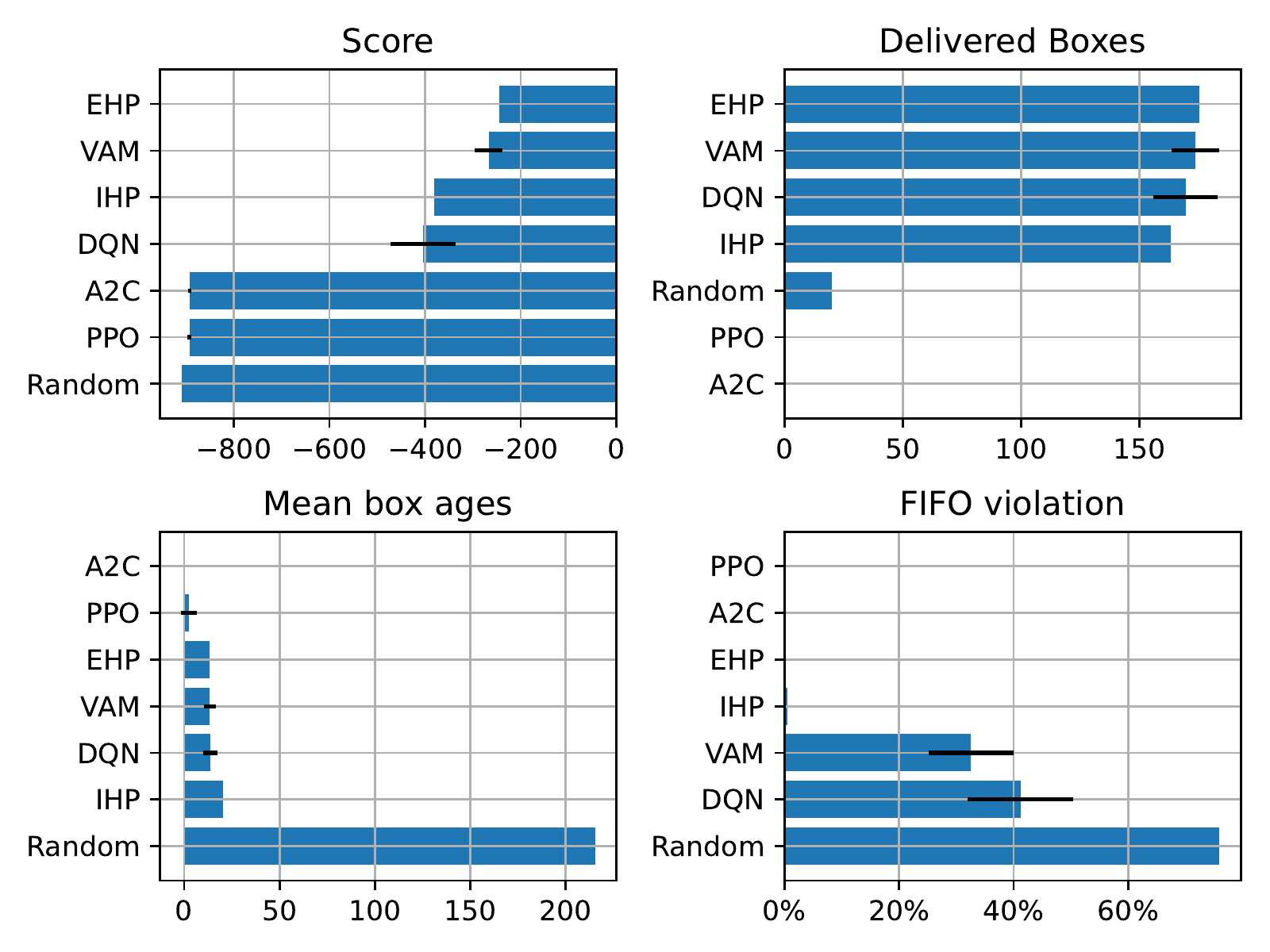}
        \caption{Environment metric comparison for the defined policies.}\label{fig:un_met}
    \end{figure}

    Figure \ref{fig:un_met} displays other metrics of interest for the environment. These results are obtained by averaging the last $100$ episodes, assuming stability in the results. 
    In terms of Delivered Boxes, the results are similar to the score, but DQN manages to deliver more boxes than IHP. This might be because DQN was able to learn strategies that can improve its productivity in terms of delivered boxes, strategies that we applied to build the EHP, in fact. However, it scores worse than IHP because, even though it manages to deliver more boxes, it performs other sub-optimal actions. On the other hand, neither PPO nor A2C managed to deliver any boxes, which is evident by looking at their score. The consequence of this fact is that the other metrics (FIFO violation and Mean Box ages) show apparently good results, but they are not feasible because these metrics are not valid for the low number of delivered material. Thus, moving on to these metrics, we have the Mean Box Ages, in which the DQN policies get very close to the EHP, enhancing the IHP. This is plausible as DQN does not have the restrictions that IHP has, and it is capable of delivering items directly from the entry points. Furthermore, EHP does not have these restrictions either, so it is able to further reduce the mean age. In addition, the FIFO Violation metric shows that DQN is able to build those solid numbers by not delivering the oldest item $40\%$ of the time, while the VAM policy enhances this metric, achieving a violation rate of $30\%$. As a side note, the small percentage of FIFO violation by the IHP policy is due to the fact that, in this policy, the occupancy of the warehouse is nearly at its maximum and, as a consequence, restricted cells start to appear. Consequently, in a small number of cases, the oldest boxes of the warehouses are in a restricted position and, therefore, the IHP agent must deliver other boxes, violating the FIFO method.

    \subsection{Hyperparameter Optimization}

        The above results show that, even though some RL policies achieve near-human performance, others have ample room for improvement. This is due to the fact that, for the training of these policies, we used the default hyperparameters from stable-baselines3. In this section, we performed the hyperparameter optimization as described in Section~\ref{sec:hyperp}.

        \begin{figure}[t]
            \centering
            \includegraphics[width=0.487\textwidth]{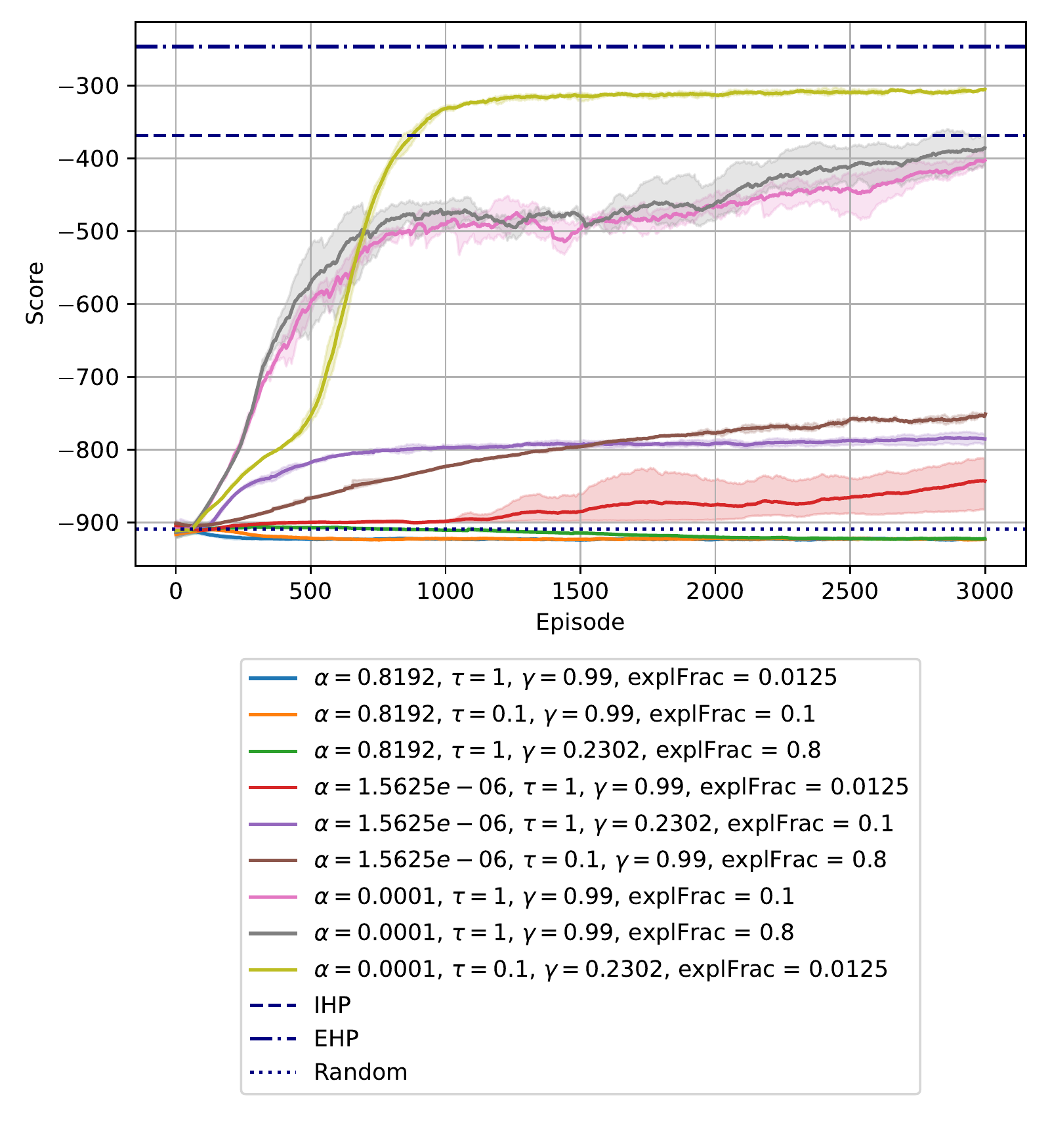}
            \caption{Hyperparameter tuning process for the DQN policy over 4 runs.}\label{fig:doe}
        \end{figure}
        
         Figure \ref{fig:doe} shows the result of the tuning process for DQN. We observe that some hyperparameter combinations cannot even compete with the random policy, while other combinations can even compete with the IHP, although being quite far away from the EHP yet. Finally, we have a combination that enhances the IHP and approaches the EHP, the yellow line with the hyperparameter combination of $\alpha = 0.001$, $\tau = 0.1$, $\gamma = 0.23$ and $\text{explFrac} = 0.0125$. As a reference, the pink line (with $\alpha = 0.001$, $\tau = 1$, $\gamma = 0.99$ and $\text{explFrac} = 0.1$) corresponds to the default hyperparameters of stable-baselines3. We can conclude that these original hyperparameters were a good combination, but we were able to enhance performance by choosing a more suitable combination.
        
        We performed a similar analysis with the other RL algorithms, namely A2C and PPO. Table \ref{tab:hp_comb} shows the combinations and we highlighted the best combinations for each case.
        \begin{table}[]
            \centering
            \resizebox{0.35\textwidth}{!}{%
            \begin{tabular}{ccccc}
            & \multicolumn{1}{l}{$\alpha$} & \multicolumn{1}{l}{$\gamma$} & \multicolumn{1}{l}{vfCoef} & \multicolumn{1}{l}{$\backepsilon$} \\ \hline
            \multirow{9}{*}{\textbf{A2C}}  & 1.5625e-06      & 0.99            & 0.1          & -    \\
                                        & 1.5625e-06      & 0.2302          & 0.9          & -    \\
                                        & 1.5625e-06      & 0.2302          & 0.5          & -    \\
                                        & 0.8192          & 0.99            & 0.9          & -    \\
                                        & 0.0001          & 0.2302          & 0.9          & -    \\
                                        & 0.8192          & 0.2302          & 0.5          & -    \\
                                        & 0.0001          & 0.99            & 0.5          & -    \\
                                        & 0.8192          & 0.2302          & 0.1          & -    \\
                                        & \textbf{0.0001} & \textbf{0.2302} & \textbf{0.1} & -    \\ \hline
            \multirow{11}{*}{\textbf{PPO}} & 1.5625e-06      & 0.99            & 0.1          & 0.08 \\
                                        & 1.5625e-06      & 0.99            & 0.9          & 0.6  \\
                                        & 0.0001          & 0.99            & 0.9          & 0.2  \\
                                        & 0.8192          & 0.2302          & 0.9          & 0.08 \\
                                        & 0.8192          & 0.99            & 0.1          & 0.6  \\
                                        & 1.5625e-06      & 0.99            & 0.5          & 0.6  \\
                                        & 0.8192          & 0.99            & 0.5          & 0.2  \\
                                        & 0.0001          & 0.2302          & 0.1          & 0.08 \\
                                        & 1.5625e-06      & 0.2302          & 0.1          & 0.2  \\
                                        & 0.0001          & 0.2302          & 0.5          & 0.6  \\
            & \textbf{1.5625e-06}          & \textbf{0.99}                & \textbf{0.5}               & \textbf{0.08}                     
            \end{tabular}%
            }
            \caption{ Studied hyperparameter combinations for A2C and PPO}\label{tab:hp_comb}
            \end{table}

    \subsection{Results with Hyperparameter Optimization}

         We re-trained the RL policies using the optimal hyperparameter combination of the previous section with DQN, VAM,  A2C, and PPO.

        \begin{figure}[t]
            \centering
            \includegraphics[width=0.487\textwidth]{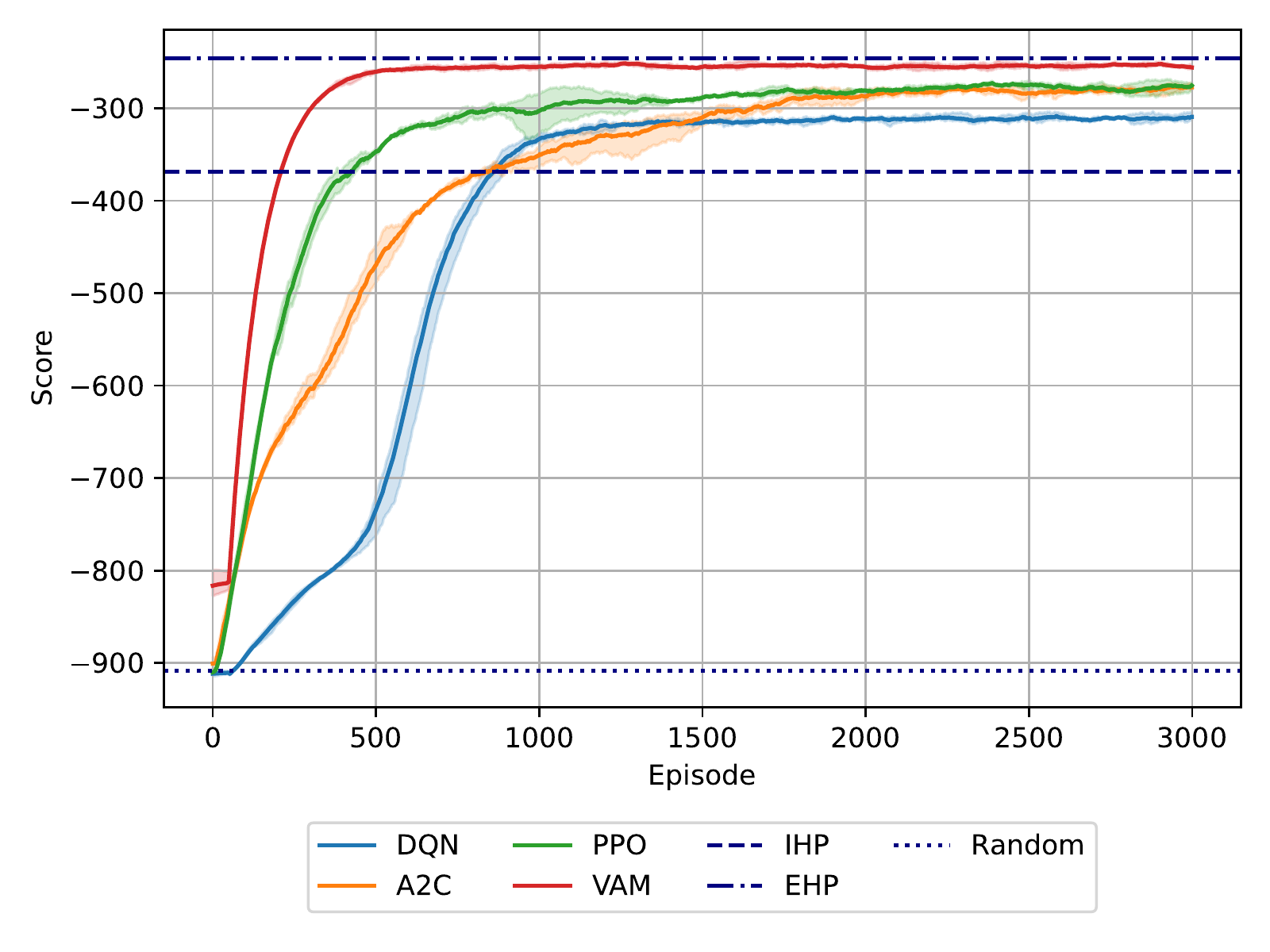}
            \caption{Score as a function of the number of collected episodes for the tested RL algorithms and baseline policies after hyperparameter optimization over 10 runs).}\label{fig:tun_algo}
        \end{figure}

        Figure \ref{fig:tun_algo} depicts the result. We note that the results are drastically enhanced, showing that policies that previously reached near-random results now perform better than the DQN policy and even the IHP, getting closer to the VAM policy and to the EHP. Even the VAM policy improves its convergence speed, although its score is quite difficult to enhance. We can conclude that, optimizing the hyperparameters, some results can be significantly improved, as now all the RL policies are better than the IHP, which at first was only surpassed by the results of the VAM policy.

        \begin{figure}[t]
            \centering
            \includegraphics[width=0.487\textwidth]{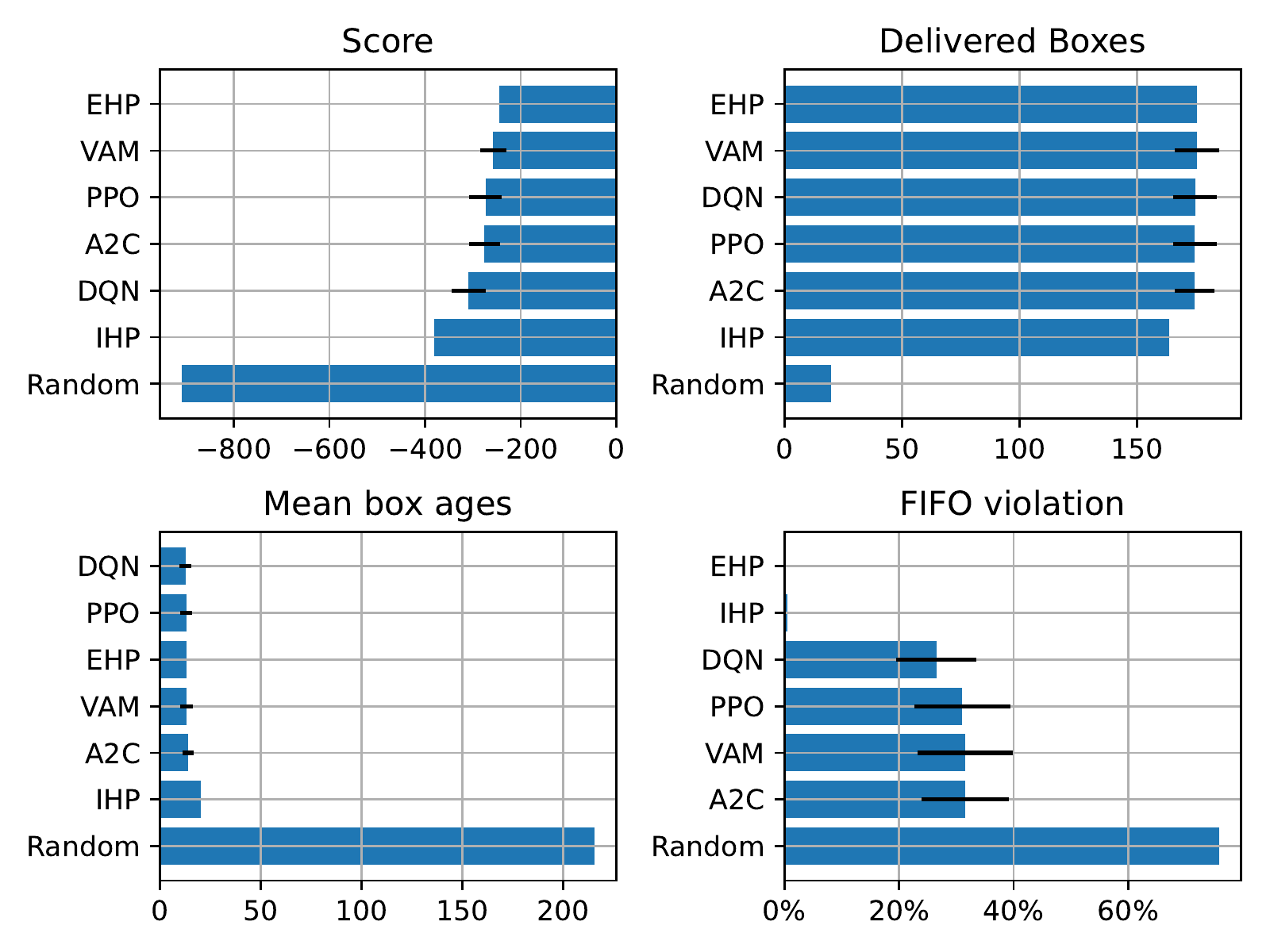}
            \caption{Metric comparison for the defined policies with optimized hyperparameters.}\label{fig:tun_met}
        \end{figure}

        As for the metrics, we can observe that the Score metric has improved, since all the RL metrics are better than the IHP, although not as good as the EHP. In terms of Delivered Boxes, each RL policy is close to the maximum score achieved by the EHP, with small divergences due to the stochasticity of the environment. The Mean Box Age metric results are also quite enhanced and it can be seen that some RL policies minimize the box ages, even more than the EHP, probably due to the fact that those policies are allowed to violate the FIFO criterion while the human policies are not.

\section{Discussion}

    In the previous section, we showed promising results. However, the \textit{Storehouse} environment shows some potential limitations, especially when it comes to scalability.
    The previous results were given on the \textit{$6 \times 6$} configuration, and that leads to a moderately large number of possible states.
    However, real warehouses tend to be much larger and, thus, the state space would increase significantly. This could be troublesome as the complexity of the problem inherently becomes greater (for example, a $100 \times 100$ grid with 25 different types of materials). This training problem impacts the hyperparameter tuning process. This process, although it was drastically reduced by using the DoE method, is slow, since several runs must be performed to find the optimal combinations, and with larger state spaces, this training would be even slower. 

    The impact of dealing with larger state spaces is unclear. We would, most likely, have similar results in terms of scoring, but the training session would last a lot longer. With the $6 \times 6$ environment, the training took around $3.5$ hours, we suspect this number may be significantly increased. However, the VAM policy would probably be able to maintain its convergence speed, since the action space varies depending on the available actions. Nevertheless, the use of more case-specific algorithms could overcome this limitation and, thus, lessen the impact of this limitation. An example of possible algorithms that would be immune to the curse of dimensionality is AlphaZero \cite{Silver2018} or adding the Hindsight Experience Replay (HER) technique \cite{Andrychowicz2017} to an already stated algorithm. Regarding hyperparameter tuning, the optimization of these parameters would help to improve the convergence speed, but we already stated that the time required for this process is quite high. However, this tuning process is the last step in an RL training, so the impact of this limitation on the general scope of the project is minimal.


%

\section{Conclusions}


    In this contribution, we covered the definition and implementation of a warehouse management solution for the efficiency problem in the logistics industry. We described the current methodologies used to solve this problem and found an improvement opportunity and proposed an RL-based solution that allows users to simulate a customizable warehouse that can be used to represent real warehouses. We also proved the feasibility of this environment by training several policies, and, after comparing the results with human-defined policies and a random policy, we saw that these policies are able to solve the problem with near-optimal performance.
    
    Therefore, we can conclude that the \textit{Storehouse} environment is a novel tool that can be used for training policies of warehouse management systems with customizable layouts and settings. This environment uses the gym interface, so it is standardized to be used with almost any RL algorithm. We also saw that, by training RL policies based on State-of-the-Art algorithms, this environment is robust and these policies are able to converge to promising results, and near-optimal performance. 

    In future scenarios, we plan to use this environment to test planning approaches and algorithms like AlphaZero. We expect that these approaches would maintain good results even with larger state spaces. 
    
%


\bibliographystyle{alpha}
\bibliography{sample}

\end{document}